\documentclass[times,twocolumn,final,authoryear,preprint]{elsarticle}

\usepackage{ycviu}
\usepackage{framed,multirow}

\usepackage{amssymb}
\usepackage{latexsym}

\usepackage{url}
\usepackage{xcolor}
\definecolor{newcolor}{rgb}{.8,.349,.1}

\usepackage{etex}
\usepackage{amssymb,amsmath,epsfig}
\usepackage{pifont}
\usepackage{amsmath}
\usepackage{multirow}
\usepackage{textcomp}
\usepackage{booktabs}
\usepackage{graphics}
\usepackage{diagbox}
\usepackage{tabularx}
\usepackage{arydshln}
\usepackage{array}
\usepackage{makecell}
\usepackage{pgfplots}
\usepackage{bbm}
\usetikzlibrary{positioning}	
\usepgfplotslibrary{patchplots}
\DeclareMathOperator*{\argmax}{argmax}

\newcommand{\setvariable}[2]{
    \let#1\relax
    \newcommand{#1}{#2}
}

\journal{Computer Vision and Image Understanding}

\begin{document}

\thispagestyle{empty}

\begin{frontmatter}

\title{Pushing the boundaries of audiovisual word recognition using Residual Networks and LSTMs}

\author[1]{Themos \snm{Stafylakis}\corref{cor1}} 
\cortext[cor1]{Corresponding author:}
\ead{themosst@gmail.com}
\author[1,2]{Muhammad Haris \snm{Khan}}
\author[1]{Georgios \snm{Tzimiropoulos}}

\address[1]{Computer Vision Laboratory, University of Nottingham, UK}
\address[2]{Electrical Engineering Department, COMSATS Lahore Campus, Pakistan}

\received{1 May 2013}
\finalform{10 May 2013}
\accepted{13 May 2013}
\availableonline{15 May 2013}
\communicated{S. Sarkar}

\begin{abstract}
Visual and audiovisual speech recognition are witnessing a renaissance which is largely due to the advent of deep learning methods. In this paper, we present a deep learning architecture for lipreading and audiovisual word recognition, which combines Residual Networks equipped with spatiotemporal input layers and Bidirectional LSTMs. The lipreading architecture attains 11.92\% misclassification rate on the challenging Lipreading-In-The-Wild database, which is composed of excerpts from BBC-TV, each containing one of the 500 target words. Audiovisual experiments are performed using both intermediate and late integration, as well as several types and levels of environmental noise, and notable improvements over the audio-only network are reported, even in the case of clean speech. A further analysis on the utility of target word boundaries is provided, as well as on the capacity of the network in modeling the linguistic context of the target word. Finally, we examine difficult word pairs and discuss how visual information helps towards attaining higher recognition accuracy.
\end{abstract}


\end{frontmatter}


\section{Introduction}
Audiovisual automatic speech recognition (ASR) is the problem of recognizing speech by combining audio and video information. It is well known that auditory perception can be profoundly modified by visual information for the speaker's lip movements\footnote{Accepted to {\bf Computer Vision and Image Understanding (Elsevier)}. DOI: \url{https://doi.org/10.1016/j.cviu.2018.10.003}} \citep{mcgurk1976hearing,macdonald1978visual}. 

In the past decades, several approaches for integrating visual information in ASR systems were introduced. Most of them were based on encoding visual information with handcrafted features or statistical shape models (e.g. active shape models) and modeling the sequence of hidden states and the emission probability of each state \citep{potamianos2003recent} with continuous density hidden-Markov models (HMMs). Despite their limitations, such as the controlled environments and the poor generalizability to speakers unseen during training, those systems demonstrated the efficacy of visual information in improving the ASR accuracy especially in noisy environments. More recently, the tremendous progress witnessed in audio-based ASR and computer vision due to deep learning methods, together with the release of large, publicly available and ``in-the-wild'' audiovisual databases have refreshed the interest in audiovisual ASR. Deep learning methods originally introduced in audio-only ASR are combined with computer vision models (e.g. convolutional neural networks, CNNs) and achieve vast improvements in recognition accuracy even in unconstrained videos \citep{chung2017lipsent}.   

In this article, we introduce an audiovisual architecture which goes beyond state-of-the-art in visual and audiovisual word recognition. We focus on the Lipreading-In-The-Wild (LRW) database, which is the largest publicly available database for the task \citep{chung2016lip, chung2018learning}. The evaluation protocol is closed-set word identification, where each video contains one of the 500 target words. However, target words are not isolated, but appear within phrases, which are excerpted from BBC-TV and are of fixed duration (1.16s). Moreover, word boundaries indicating the beginning and the end of each target word are given \citep{chung2016lip}.

Our lipreading architecture is composed of a Residual Network (ResNet) with spatiotemporal (3D) input layers and a Bidirectional LSTM (BiLSTM) backend. We explore several configurations, the best of which attains 11.92\% misclassification rate (MCR). Our audio architecture is based entirely on BiLSTM, it uses plain log-spectral features and it is designed in such a way so that it can be naturally integrated with the lipreading network. Contrary to typical stacks of BiLSTMs, where the two directional outputs are concatenate after every BiLSTM layer, we concatenate them only at last BiLSTM layer. To examine the gain in performance under noisy conditions, we use the DEMAND dataset, which contains several types of environmental noise. The performance of the audiovisual system is compared to the audio-only system under different SNR levels, as well as to score-level fusion (i.e. late integration \citep{katsaggelos2015audiovisual}). Moreover, we examine word confusion between challenging word pairs and we show how visual information helps towards reducing it.

LRW is a recently released database and one of its main properties is that target words appear within utterances with given target word boundaries. Therefore, a thorough experimentation on the optimal use of target word boundaries and on the way the network may utilize the context of the target word is worth to be conducted. To this end, we first examine ways of exploiting the word boundaries and we show that passing them as indicator variables yields substantially superior results compared to using them in order to remove frames preceding or succeeding the target word (out-of-boundaries frames), as proposed in \cite{chung2018learning}. Moreover, we test whether this improvement is due to an implicit modeling of the linguistic context of the target word. By training and evaluating a network entirely on out-of-boundaries frames, we show that a small fraction of target words can be recognized using merely their context.

The rest of the paper is organized as follows. In Section \ref{Related_Work} we present some of the recent works in lipreading and audiovisual speech recognition, while in Section \ref{LRW} we give a description of the ``Lipreading-In-The-Wild'' database. In Section \ref{Lipreading} we describe our lipreading architecture while the audio and audiovisual architectures are discussed in Section \ref{Audiovisual}. In Section \ref{Sub:LRE} we present experiments using several lipreading architectures and configurations, followed by audio and audiovisual experiments in Section \ref{Sub:AVE}. In Section \ref{Sect:EA} we provide a further analysis on the results, e.g. by focusing on difficult word pairs and trying to quantify the contribution of the word context in the results. Finally, conclusions and further applications of the proposed ResNet features and training scheme are discussed in Section \ref{Sect:Conc}.  

\section{Related work}
\label{Related_Work}
Research on visual speech recognition witnessed a first wave of interest few decades ago, either as a stand-alone application or as a means to enhance audio-only ASR \citep{zhou2014review}. Traditional visual ASR systems typically deploy an appearance or shape-based feature extractor, a set of linear or nonlinear transforms for suppressing the undesired variability associated to speaker, pose, or other sources, and an HMM with which transitions between visemes are modeled. Visemes are defined as the set of static or dynamic mouth shapes representing clusters of contrastive phonemes (e.g. /p, b, m/, and /f, v/) \citep{taylor2012dynamic}. 
Appearance-based visual features are typically derived by applying PCA, discrete cosine transform (DCT), or discrete wavelet transform (DWT) on the region of interest, while shape-based features model the contours of the speaker’s lips, via geometric descriptors or statistical models of shape, such as active shape models \citep{dalton1996automatic, luettin1997speechreading,chan2001hmm}. Common ASR transforms such as Linear Discriminant Analysis (LDA) and feature-space Maximum Likelihood Linear Regression (fMLLR) are also applied to visual features, in order to enhance their discriminability and to adapt them to the speaker's characteristics \citep{potamianos2003audio}. Active appearance models (AAM), i.e. generative models for the shape and the texture of faces (and objects in general) have also been very effective in visual and audiovisual ASR. Apart from their appealing properties as features which combine shape and texture information, they offer a principled way for dynamically estimating the visual feature observation uncertainty, allowing for uncertainly-aware HMM training and fully adaptive audiovisual integration schemes \citep{papandreou2009adaptive,matthews2002extraction}. 

Over the past few years, a new wave of visual and audiovisual ASR architectures has been observed, which are largely inspired by advances in deep learning and especially those emerged in audio-only ASR and computer vision. The use of Recurrent Neural Networks (RNNs) was first explored in \cite{wand2016lipreading}, where Long Short-Term Memory (LSTM) networks are applied to the GRID corpus. The proposed network is speaker dependent, as a result of the small speaker variability in GRID which makes generalizability to new speakers hard. A speaker independent extension of the network is proposed in \cite{Wand2017}, where features are mapped to a common speaker independent space using domain adversarial neural networks \citep{ganin2015unsupervised}. In \cite{assael2016lipnet} LipNet is introduced, a system which combines Convolutional Neural Networks (CNNs) and Gated Recurrent Units (GRUs), capable of performing sentence-level recognition on GRID. The lack of frame-level annotation is addressed by Connectionist Temporal Classification (CTC, \cite{graves2006connectionist}), which together with attentive sequence-to-sequence neural networks are the two dominant approaches in end-to-end ASR \citep{Takaaki2017}. LipNet is also the first lipreading model to employ spatiotemporal convolutions in the input layers. In \cite{Potamianos2017} the authors propose the use of convolutional LSTMs with skip connections and perform experiments on the IBM audiovisual database of connected digits. The proposed architecture combines the strength of LSTMs in sequence modeling with the efficacy of spatial convolutions in a unified model. The authors report drastic improvements over conventional GMM-HMM based systems with handcrafted features \citep{potamianos2003audio}. In \cite{chung2018learning}, the research team which collected ``Lipreading-In-The-Wild'' improves its previous architecture \citep{chung2016lip} by combining a VGG network with an LSTM backend. The improved architecture attains 36.0\% WER, which is further reduced to 29.5\% when the target word boundaries are used to discard out-of-boundaries frames. Very recently, an improvement of the LipNet is introduced in \cite{shillingford2018large}. A deep architecture comprising spatiotemporal convolutions, BiLSTMs, phoneme-distributions and CTC loss is trained on a new large-scale database for lipreading. The authors report 40.9\% WER, outperforming LipNet and attentive encoder-decoder architectures \citep{chung2017lipsent} by a large margin. Finally, in \cite{afouras2018deep}, the authors propose a lipreading system deploying self-attention layers (Transformer network) \citep{vaswani2017attention}. The experiments show that it yields better results compared to architectures with convolutional and BiLSTM-based backends.

In parallel, research in integrated audiovisual ASR architectures is moving towards deep learning end-to-end methods. In audiovisual ASR, intermediate and late integration (or fusion) are the two main approaches deployed to combine the two modalities. Intermediate integration refers to those approaches having modality-specific front-ends, with the two modalities being merged in the backend of the audiovisual architecture. On the other hand, in late integration, two separate systems are trained and their probabilistic outputs are combined, by applying a weighted average over the logarithmic likelihoods or posterior probabilities \citep{katsaggelos2015audiovisual}. Hybrid approaches exist as well, such as architectures where the two modalities are merged at the decision level with a bilinear layer, enabling joint bi-modal training \citep{mroueh2015deep}, or approaches using gating layer to reduce the effect of noisy or uninformative visual features \citep{tao2018gating}. In \cite{petridis2017end}, the authors propose an end-to-end intermediate integration architecture with fully connected layers in the frontend and a single-layer BiLSTM in the backend. The results reported on OuluVS2 database demonstrate a notable improvement over the audio-only version of the architecture, when the Signal-to-Noise Ratio (SNR) is $\leq$15dB \citep{anina2015ouluvs2}. In \cite{mroueh2015deep}, audiovisual architectures with fully connected layers in the frontend and three different integration approaches are compared. The first architecture is a intermediate integration approach, where the two modalities are concatenated in the feature space and passed as input to a Deep Neural Network (DNN). The second is a late fusion approach, where the scores are averaged on the log-posterior domain, while the third is a hybrid approach where merging is performed by a factored bilinear SoftMax layer. The experimental results performed on the IBM AV-ASR Large Vocabulary Studio Dataset show that the first two approaches perform equally well and clearly better that the third one. Finally, in \cite{chung2017lipsent}, an end-to-end audiovisual architecture is proposed, which expands the ``Listen, Attend and Spell'' audio-only ASR approach to visual and audiovisual ASR \citep{chan2016listen}. The architecture is based on sequence-to-sequence models (also known as encoder-decoder) with attention. Intermediate integration is performed by the speller (i.e. the decoder), whose estimates are conditioned on two context vectors, one for each modality. The system is evaluated on the Lipreading Sentences in-the-wild (LRS) database, which is the second large and in-the-wild audiovisual database made available recently \citep{chung2017lipsent}. Apart from the experiments on LRS, the authors conduct experiments with a lipreading version of their architecture on GRID (attaining the best published results) as well as on LRW.

\section{The Lipreading-In-The-Wild database}
\label{LRW}
LRW is the largest publicly available audiovisual database for word-level audiovisual speech recognition. It is composed of short videos from BBC news and talk-shows, excerpted using a pipeline for fully automated large-scale data collection from TV broadcasts \citep{chung2016lip}. It is characterized by high variability in speakers, motion and resolution, as well as by a large number of target words and training instances per word. Pose variability is clearly higher compared to other datasets (e.g. GRID), although views are in general between $0^{\circ}$ and $30^{\circ}$. Due to its large training set, training deep architectures from scratch is feasible, using standard data augmentation (e.g. random crops, horizontal flips) and regularization methods (e.g. dropouts, batch normalization). The main statistics and characteristics of LRW are summarized in Table \ref{LRWstats}. 

\begin{figure*}
\centering

\begin{tikzpicture}[font=\scriptsize,inner sep=0,outer sep=0]
	\node (e3r8) at (0,0) {\includegraphics[width=\subfigwidth]{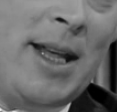}};
	\node[right=0ex of e3r8.east] (e3r9) {\includegraphics[width=\subfigwidth]{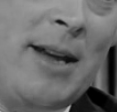}};
	\node[right=0ex of e3r9.east] (e3r10) {\includegraphics[width=\subfigwidth]{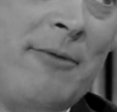}};
    \node[right=0ex of e3r10.east] (e3r11) {\includegraphics[width=\subfigwidth]{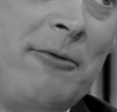}};
    \node[right=0ex of e3r11.east] (e3r12) {\includegraphics[width=\subfigwidth]{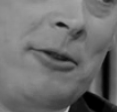}};
    \node[right=0ex of e3r12.east] (e3r13) {\includegraphics[width=\subfigwidth]{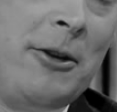}};
    \node[right=0ex of e3r13.east] (e3r14) {\includegraphics[width=\subfigwidth]{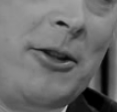}};
    \node[right=0ex of e3r14.east] (e3r15) {\includegraphics[width=\subfigwidth]{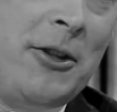}};
    \node[right=0ex of e3r15.east] (e3r16) {\includegraphics[width=\subfigwidth]{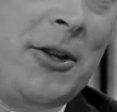}};
    \node[right=0ex of e3r16.east] (e3r17) {\includegraphics[width=\subfigwidth]{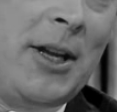}};
    \node[right=0ex of e3r17.east] (e3r18) {\includegraphics[width=\subfigwidth]{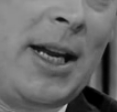}};
    \node[right=0ex of e3r18.east] (e3r19) {\includegraphics[width=\subfigwidth]{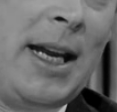}};
    \node[right=0ex of e3r19.east] (e3r20) {\includegraphics[width=\subfigwidth]{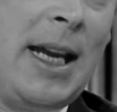}};
    \node[right=0ex of e3r20.east] (e3r21) {\includegraphics[width=\subfigwidth]{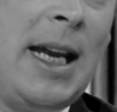}};
    \node[right=0ex of e3r21.east] (e3r22) {\includegraphics[width=\subfigwidth]{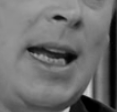}};
    \node[right=0ex of e3r22.east] (e3r23) {\includegraphics[width=\subfigwidth]{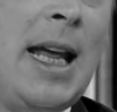}};                
\end{tikzpicture}

\caption{Frame sequence from LRW, corresponding to the word EVERYTHING. Frames are cropped as those we use as input to the lipreading network.}
\label{stream}
\end{figure*}

There are several word pairs with small pairwise edit distance, such as MILLION-BILLION, SPEND-SPENT, RUSSIA-RUSSIAN, BENEFIT-BENEFITS. Moreover, the target words occupy the central frames of the utterance, making word recognition even more challenging due to co-articulation with preceding and succeeding words. Target word boundaries are given for all splits (training, development and test) and algorithms may use them to localize the target words. Finally, the faces are registered and therefore the region of interest (mouth region) can be extracted directly using fixed cropping for all frames and videos, without applying face detection and facial landmark detection and tracking. Fig. \ref{stream} depicts a sequence of frames from an LRW video corresponding to the word EVERYTHING (frames lying out of the word boundaries are not depicted). The region of interest is defined similarly to \cite{chung2017lipsent}, i.e. it is square and relatively wide.  

\begin{table}[!htbp]
\caption{\label{LRWstats} LRW database statistics and characteristics}
\centering
\begin{tabular}{ c  c }
\Xhline{2\arrayrulewidth}  
Target words & 500 \\
\hline 
Training instances & 488766 \\
\hline 
Development instances & 25000 \\ 
\hline 
Test instances & 25000 \\ 
\hline 
Training instances per word & 800 - 1000 \\ 
\hline 
Range of views & $0^{\circ}$ - $30^{\circ}$ \\
\hline 
Video duration & 1.16s \\
\hline 
Frames per video & 29 \\
\hline 
Speakers & $>$1000 \\
\Xhline{2\arrayrulewidth} 
\end{tabular}
\end{table}
\setvariable\subfigwidth{4.5em}

\section{Lipreading Network}
\label{Lipreading}
In this section, we focus on the visual part on the network. The network can be naturally partitioned into a frontend and a backend. The frontend is a modified 18-layer Residual Network, while the backend is a 2-layer BiLSTM, modeling the probability of the input sequence containing each target word. 
\subsection{Frontend: ResNet with spatiotemporal input layers}
We use an $18$-layer ResNet, which we found to performing equally well with the 34-layer we used in \cite{Stafy2017}. ResNets are convolutional networks with skip connections, which facilitate training of deep architectures \citep{he2016identity}. They can be decomposed into 4 different blocks, each reducing the spatial size while increasing the number of channels by a factor of 2. We make two modifications on the ResNet to make it more effective for the particular task. (a) We replace the first spatial (i.e. 2D) layers with their spatiotemporal (i.e. 3D) analogues \citep{Tran2015,ji20133d}. The first convolutional layer has kernel size of $64\times5\times7\times7$ (channels$\times$time$\times$width$\times$height), while max pooling has a kernel size equal to $1\times1\times3\times3$. At this stage, we do not reduce the time resolution, therefore we set the temporal stride equal to $1$. (b) We replace the average pooling layer of the ResNet output by a fully connected layer. The average pooling layer is more adequate for problems requiring translation invariance (e.g. object recognition and detection) but it is less justified for images of the mouth region. The block of layers transform the tensors as $1\times112\times112 \rightarrow 64\times28\times28 \rightarrow 128\times14\times14 \rightarrow 256\times7\times7 \rightarrow 512 \times4\times4$ (channels$\times$width$\times$height), while the last set of layers transforms them as $512\times4\times4 \rightarrow 8192\times1\times1 \rightarrow 256\times1\times1$ (the temporal dimension is suppressed for clarity and its size is fixed, $T^V = 29$).    
\subsection{Backend I: Temporal convolutional}
The first backend we deploy is a temporal convolutional network. It is a rather simple backend, the main purpose of which is to initialize the ResNet (we did not succeed in training from scratch the architecture with BiLSTM backend). Its building blocks are composed of a temporal convolutional layer, followed by batch normalization, Rectified Linear Units (ReLU) and temporal max pooling. Each of the two layers reduces the time resolution by a factor of $4$ (by using a stride equal to $2$ in the convolutional and max pooling layers), while the number of channels is increased by $2$. The tensor sizes (time$\times$channels) are therefore $29\times256 \rightarrow 7\times512 \rightarrow 1\times1024$. The final linear layers implement a bottleneck $1024 \rightarrow 256 \rightarrow 500$, followed by a SoftMax layer to approximate the posterior probability over the $N_{w} = 500$ target words. 
\subsection{Backend II: Bidirectional LSTMs}
Our proposed backend is composed of a two-layer BiLSTM (without peepholes) of $256$-size, followed by an average pooling layer which aggregates information across time-steps, a fully connected (FC) linear and a SoftMax layer. Our two-layer BiLSTM differs from the usual stack of two BiLSTM layers; we obtained significantly better results by concatenating the two directional outputs only at the output of the second LSTMs. The backend receives as input the collection of $256$-size features extracted by the ResNet, augmented by a binary variable indicating whether or not the frame lies inside or outside the word boundaries, i.e.

\begin{equation}
\mathbf{x}^{+}_t = \left[ \mathbf{x}_t,b_{t} \right],
\end{equation} 
where
\begin{equation}
\mathbf{x}_t = \mbox{ResNet}(D_v)_t
\end{equation}
is the $t$-th output of the ResNet, $D_v$ is the collection of frames and
\begin{equation}
b_t = \mathbbm{1}_{ t \in B }
\end{equation}
is the indicator variable denoting whether $t$-th frame lies within the word boundaries $B$. 

The BiLSTM layers are followed by an average pooling layer, which aggregates temporal information into a fixed length representation. We apply dropouts with $p=0.30$ to the inputs of each LSTM (yet not to the recurrent layer, see \cite{LSTMdrop2017}). Following the theoretical justification in \cite{DropoutsRNN}, a single mask is sampled and applied to all frames of a given sequence. Finally, batch normalization is applied to the average pooling layer, together with a dropout layer with $p=0.15$ \citep{ioffe2015batch}.

\begin{figure}[!htbp]
\centering
\includegraphics[width=2.90in]{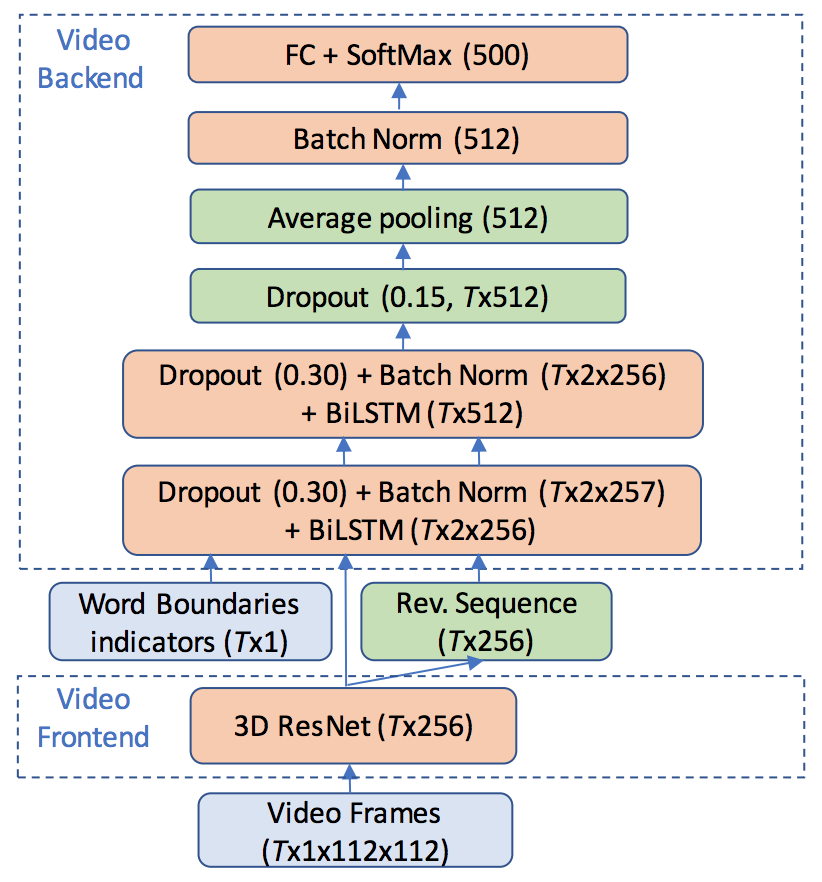}
\vspace{-2mm}\caption{The block-diagram of the lipreading network with BiLSTM backend. Tensor sizes correspond to the output tensors of each module. Inputs are depicted with blue, trainable components of the network with orange, and non-trainable components with green.}
\label{diagram}
\end{figure}

\subsection{Training algorithm, optimization and evaluation}
We train the network by using the cross entropy criterion over word labels with ADAM optimizer \citep{Adam}. The initial learning rate is equal to $3\times10^{-3}$, the final is equal to $10^{-5}$, and we reduce it by a factor of 2 when no progress is attained on the validation set after 3 consecutive epochs. 
We evaluate the performance using the argmax operator, i.e.
\begin{equation} 
\hat{w} = \argmax_w p(w|D_{v}, \Theta_{v}).
\end{equation}
where $\Theta_{v}$ is the set of parameters of the visual network, $D_{v}$ is a frame sequence, and $w \in \cal{W}$ is a word of the vocabulary of target words $\cal{W}$, where $\left|\cal{W}\right| = N_w = 500$.
 
\section{Audio and Audiovisual Networks}
\label{Audiovisual}
In this section, our architecture for audio-based speech recognition is introduced and discussed, together with intermediate and late audiovisual integration.  
\subsection{Proposed architecture for audio-only word recognition}
\begin{figure}[!htbp]
\centering
\includegraphics[width=3.30in]{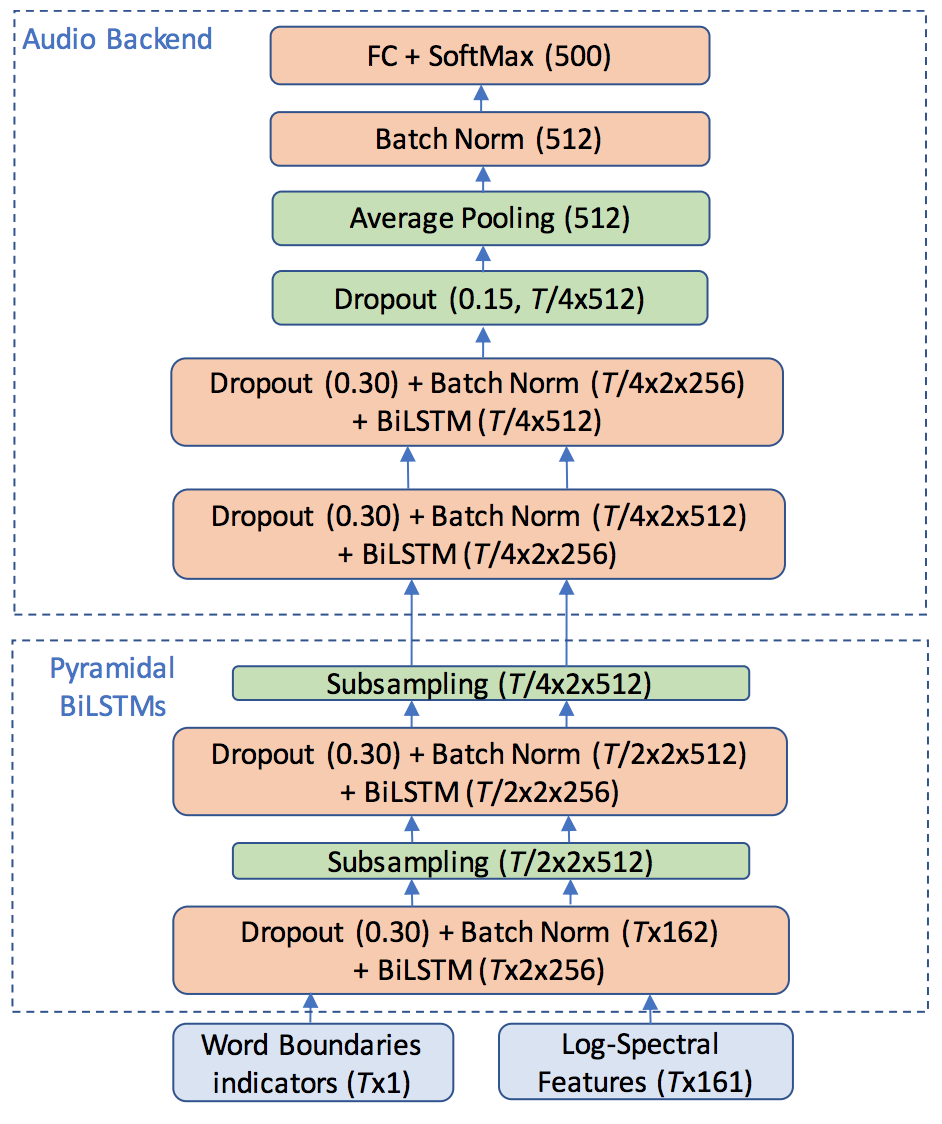}
\vspace{-2mm}\caption{The block-diagram of the audio network. Pyramidal are the two BiLSTM which are followed by a subsampling layer. $T$ corresponds to the set of audio features, i.e. $T = T^A=116$.}
\label{audio_diagram}
\end{figure}

Contrary to the visual network, the audio network is based entirely on BiLSTMs, which have demonstrated an enormous capacity in audio-only ASR \citep{sak2014long,chan2016listen}. We use a deep 4-layer BiLSTM architecture with the two directions being concatenated only at the output of the final LSTMs (similarly to the backend on the lipreading network). The audio is modeled using 161-dimensional log-spectral features, extracted at 100fps frame rate, i.e. 4 times the visual frame rate. We compensate for this difference by adding a subsampling layer to the output of the first two BiLSTMs, each dropping the frame rate by a factor of 2. As a result, the sequence of audio features which is passed to the backend has 25fps frame rate. There are at least two approaches to implement this reduction in the audio frame rate. The first is to subsample the LSTM output vectors used as input to the next layer, e.g. to pass only the even frames to the next LSTM. The second is via frame concatenation. In this latter case, we concatenate two consecutive LSTM output vectors and use them as input to the subsequent LSTM layer. This approach is often called pyramidal LSTM (pLSTM) and has shown its efficacy when combined with the attentive encoder-decoder architecture for audio-only ASR introduced in \cite{chan2016listen}. The standard LSTM equation (ignoring the internal mechanism) is as follows
\begin{equation}
\mathbf{h}_{t}^{l} = \mbox{LSTM}(\mathbf{h}_{t-1}^{l},\mathbf{h}_{t}^{l-1}),
\end{equation} 
where $l$ and $t$ denote LSTM layer and time index, respectively, while $\mathbf{h}_{t}^{l}$ is the output vector of the LSTM. The pLSTM is as follows
\begin{equation}
\label{pLSTM}
\mathbf{h}_{t}^{l} = \mbox{pLSTM}(\mathbf{h}_{t-1}^{l},\left[\mathbf{h}_{2 t}^{l-1},\mathbf{h}_{2t+1}^{l-1} \right]),
\end{equation} 
where $\left[ \cdot,\cdot \right]$ denotes vector concatenation.
Similarly to the lipreading network, word boundaries are passed to the network as a sequence of binary indicator variables, which are concatenated with the spectral features. We apply batch normalization to the input of each LSTM as well as dropouts. The architecture is illustrated in Fig. \ref{audio_diagram}. We denote by $T^A$ the number of audio input frames, where $T^A = 4\times T^V = 116$ in LRW.

Audio features are extracted every 10ms with a sliding window of 20ms. The resulting 161-sized features are normalized with respect to mean and variance, with statistics estimated from the whole utterance. Note that the statistics are common to all frequency bins, i.e. a scalar mean and variance is estimated from and applied to all 161 bins. Estimating robust statistics for each frequency bin is hard for two reasons. First, the duration of the utterances is very short (1.16sec). Second, we use plain log-spectral features, and therefore we do not average over neighboring frequency bins as in the case of MFCC and filterbank features.

\subsection{Audiovisual architectures and integration}
\begin{figure}[!htbp]
\centering
\includegraphics[width=2.90in]{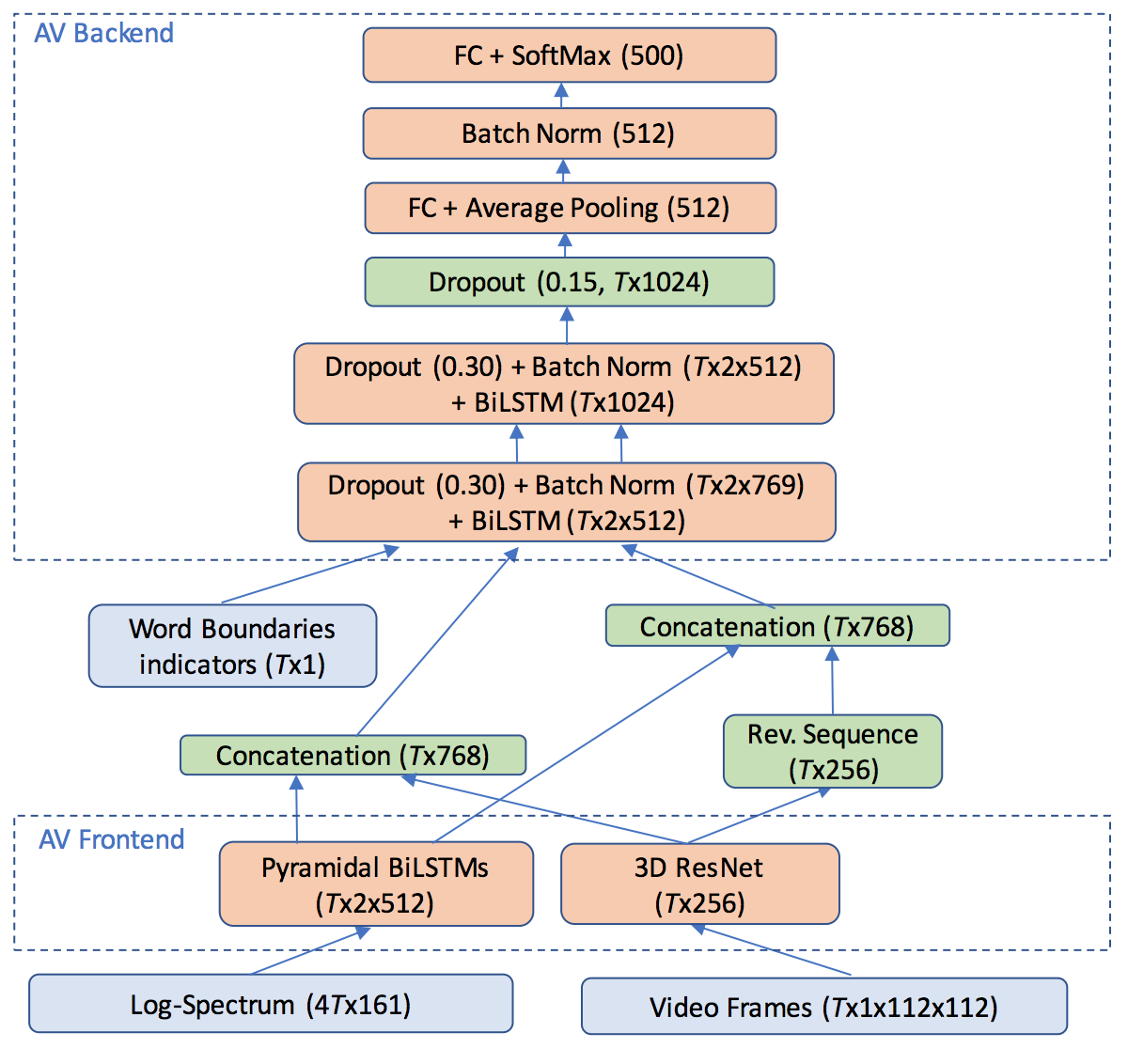}
\vspace{-2mm}\caption{The block-diagram of the audiovisual network. $T$ corresponds to the set of video frames, i.e. $T = T^V=29$.}
\label{AV_diagram}
\end{figure}

Our first audiovisual architecture follows intermediate integration and is derived by merging the audio and lipreading networks in the backend. As Fig. \ref{diagram} and Fig. \ref{audio_diagram} show, the two networks share the same backend and frame rate, and we can therefore unify them by concatenating the frontend outputs. The way we concatenate the audio with the visual frontend is shown in the lower part of Fig. \ref{AV_diagram}. We observe that audio, visual and word boundaries sequences are concatenated separately in each of the two directions, while the concatenation is taking place at the output of the two front-ends. This is slightly different compared to the audio-only network, where word boundaries sequences are concatenated directly with spectral features sequences.

For regularization, we deploy a version of multimodal training method suggested in \cite{chung2017lipsent}. More specifically, we randomly drop either the audio or the visual frame sequence, each with probability $0.25$, i.e. $p_{d}^{A}=p_{d}^{V}=0.25$. Moreover, we drop the word boundaries indicator variables with probability $p_{d}^{W}=0.25$. Multimodal training is a special case of dropout regularization, where a single binary variable is sampled for each sequence of audio, video and word boundaries indicator variables. Moreover, sampling is dependent for the two modalities, so that the probability of dropping both modalities is equal to 0, while the probability of keeping both modalities is equal to $0.50$.

Note that due to the lower frame rate and the spatiotemporal input layers, visual features span a wide context window. More specifically the visual context window is equal to $5\times40$ms = $200$ms, where 5 corresponds to the temporal size of the ResNet. Hence, for the concatenation to be meaningful, the acoustic features should have a wide context window, too. The 2-layer pyramidal BiLSTM (pBiLSTM) acoustic frontend is by construction capable of extracting features spanning wide time windows. Therefore, apart from the requirement of reducing the audio frame rate by a factor of 4, the further requirement of extracting acoustic features with sufficiently wide context window is a further rationale for deploying the particular 2-layer pBiLSTM acoustic frontend.    

Apart from intermediate integration, we also experiment with late integration. In this case, two unimodal systems are trained independently and the output of the audiovisual system is calculated as the weighted sum of the log-posteriors, which is equivalent to 

\begin{equation} 
p(w|D_{a},D_{v},\Theta_{a},\Theta_{v}) \propto p(w|D_{v},\Theta_{v})^\gamma p(w|D_{a},\Theta_{a})^{1 - \gamma},
\end{equation}
and the decision is based on
\begin{equation} 
\hat{w} = \argmax_w p(w|D_{a},D_{v},\Theta_{a},\Theta_{v}).
\end{equation}
The weight of the visual system is $\gamma = 0.40$, which we optimized on a noisy version of the validation set. 

\section{Lipreading experiments}
\label{Sub:LRE}
In this section, we report experimental results with our lipreading system. We start by presenting the baseline and the current state-of-the-art on LRW, followed by experiments using the temporal convolutional backend and optical flow. Results using the proposed system will follow, together with different configurations of the LSTM backend. Finally, the effectiveness of the proposed way of using the word boundaries will be examined and compared to the conventional approach of out-of-boundaries frame removal. 
\subsection{Baseline and state-of-the-art}
As baseline system we consider the architecture introduced in \cite{chung2017lipsent} called ``Watch, Attend and Spell'' (WAS). It is essentially an encoder-decoder network with temporal attention, using characters as recognition units, and it is a visual extension of the ``Listen, Attend and Spell'' audio-only ASR network introduced in \cite{chan2016listen}. The architecture is capable of performing sentence-level recognition, scoring 50.2\% Word Error Rate (WER) on the ``Lipreading Sentences in-the-wild" database. It is worth-mentioning that a professional lipreader scored 73.8\% on the same dataset, underlining the efficacy of the proposed architecture and training algorithm. The architecture was fine-tuned on the LRW training set and evaluated on its test set yielding $23.80\%$ MCR (Table \ref{BL}, WAS). The state-of-the-art in LRW is introduced by our research team in \citep{Stafy2017} and its differences with the proposed architecture have been discussed in Section \ref{Lipreading}. The network attains 17.03\% MCR in LRW (Table \ref{BL}, ResNet-BiLSTM).

\begin{table}[!htbp]
\centering
\caption{Lipreading baseline and state-of-the-art results on LRW. VGG-M/LSTM is reported in \cite{chung2018learning}, Watch, Attend and Spell in \cite{chung2017lipsent}, while ResNet/BiLSTM in \cite{Stafy2017}.}
\begin{tabular}{ c  c } 
\hline
\bf{System} & \bf{MCR (\%)} \\ [0.5ex] 
\Xhline{2\arrayrulewidth} 
VGG-M/LSTM & 28.50 \\ 
\hline
Watch, Attend and Spell & 23.80\\
\hline
ResNet/BiLSTM & 17.03 \\
\hline
\end{tabular}
\label{BL}
\vspace{1mm}
\end{table}

\subsection{Lipreading architectures: frontend, ResNet and backend}
In the first set of experiments, we present a set of results which are useful in assessing the contribution of each module. We begin with the temporal convolutional backend and we compare a network having spatial (i.e. 2D) or spatiotemporal (i.e. 3D) input layers (denoted by V1 and V2 in Table \ref{LRexp0}). Clearly, the contribution of the 3D input layers is significant, yielding 4.66\% absolute improvement. Note that the optimal temporal kernel size (equal to 5) is determined after experimentation, where we observed no gains by increasing it further. Similar conclusions about the efficacy of a spatiotemporal input layers can be found in \cite{assael2016lipnet}, where they are combined with GRUs in the backend. To determine the size of the ResNet we include an experiment (denoted by V3) using a 34-layer ResNet, as we suggested in our previous work \citep{Stafy2017}. Comparing V2 and V3 we conclude that 18-layer ResNet yields better results and it is the one we proceed with, given also its fewer parameters (about 2/3 of those in 34-ResNet). In the next experiment, we aim to assess the contribution of the fully connected layer over the baseline average pooling. We argue that average pooling is more relevant to domains where invariance towards translations is required (e.g. object recognition and detection), but it is less justifiable in the domain we examined, where the mouth region is aligned at each frame. The architecture where average pooling is replaced by a fully connected linear layer is denoted by V4. We observe a notable improvement, which demonstrates the effectiveness of a learnable layer over a fixed spatial averaging. The best results in this set of experiments are obtained when BiLSTMs are used in the backend. Network V6 combines a spatiotemporal 18-layer ResNet with BiLSTM backend and attains 17.01\% MCR. 

A question worth-posing is whether the spatiotemporal input layers are still required even when BiLSTMs are deployed. To address it, we train a network with 2D ResNet and BiLSTMs (denoted by V5). By comparing its performance to that of V6 we observe that the spatiotemporal input layers are highly beneficial even with BiLSTM backend, yielding 6.27\% absolute improvement over V5. In other words, spatial ResNet features, even when combined with powerful RNN models are incapable of competing with spatiotemporal ResNet features in modeling the motion patterns of the mouth region. The reason is that temporal correlations should be modeled over fine-grained spatial information, and the latter is not preserved in spatial ResNet features\footnote{Although the results demonstrate the efficacy of BiLSTMs, they do not necessarily reflect the efficacy of temporal convolutions. The temporal convolutional backend is kept intentionally simple, as its main purpose is to initialize the ResNet.}.

\begin{table}[!htbp]
\caption{Lipreading results with various network configurations. Comparisons: 2D vs 3D: spatial vs spatiotemporal layers in the frontend (FE), 18 vs 34: number of convolutional layers in ResNet, AP vs FC: Average pooling vs fully connected layer for eliminating the spatial dimensions in the ResNet. TConv vs BiLSTM: Temporal convolutional vs BiLSTM backend (BE).}
\centering
\begin{tabular}{ c c c c c c} 
\hline
\bf{Net} & \bf{FE} & \bf{ResNet} & \bf{Pooling} & \bf{BE} & \bf{MCR (\%)} \\ [0.5ex] 
\Xhline{2\arrayrulewidth}
V1 & 2D & 18 & AP & TConv & 29.37\\ 
\hline 
V2 & 3D & 18 & AP & TConv & 24.71\\
\hline 
V3 & 3D & 34 & AP & TConv & 25.44\\ 
\hline
V4 & 3D & 18 & FC & TConv & 22.51\\ 
\hline
V5 & 2D & 18 & FC & BiLSTM & 23.28\\
\hline
V6 & 3D & 18 & FC & BiLSTM & 17.01\\ 
\hline
\end{tabular}
\vspace{1mm}
\label{LRexp0}
\end{table} 

\subsection{Lipreading architectures with optical flow}
In this section we investigate the use of optical flow as input feature, either alone or together with frames. Optical flow is a two-dimensional vector field indicating the motion of each individual pixel between two successive frames \citep{black1993framework}. It has been extensively used in action recognition as well as in lipreading, usually combined with frames \citep{simonyan2014two,shiraishi2015optical}. To experiment with it, we extract the optical flow for each frame and we treat each of its two dimensions as a channel, either alone (i.e. $2\times112\times112$ input tensor size) or together with the (grayscale) frame (i.e. $3\times112\times112$ input tensor size). We perform experiments with the ResNet-18 and the temporal convolutional backend. The results are given in Table \ref{flow}. Interestingly, none of the four configurations yield superior results compared to using frames alone. The architectures using spatiotemporal input layers (V8 and V10) perform again better compared to those using spatial ones (V7 and V9). Although other configurations may be considered for using optical flow (e.g. by using intermediate integration \citep{simonyan2014two}), we conclude that the spatiotemporal input layers operating on frames can model temporal information sufficiently well. Therefore, for the rest of our experiments optical flow will not be included.
 
\begin{table}[!htbp]
\caption{Lipreading results with optical flow features. Comparisons: 2D vs 3D: spatial vs spatiotemporal input layers in the frontend (FE), Optical Flow vs Frames: indicated whether optical flow and regular frames are used as input.}
\centering
\begin{tabular}{ c c c c c c} 
\hline
\bf{Net} & \bf{FE} & \bf{Optical Flow} & \bf{Frames} & \bf{MCR (\%)}\\ [0.5ex] 
\Xhline{2\arrayrulewidth}
V7 & 2D & \checkmark &  & 37.79\\ 
\hline 
V8 & 3D & \checkmark &  & 34.68\\
\hline 
V9 & 2D & \checkmark & \checkmark & 39.44\\ 
\hline
V10 & 3D & \checkmark & \checkmark & 30.28\\
\hline
\end{tabular}
\vspace{1mm}
\label{flow}
\end{table} 

\subsection{Our lipreading architecture with BiLSTM backend}
For the next set of experiments we fix the frontend obtained by the V6 network and use it to extract and store ResNet features, which will be used for the rest of our experiments. Working directly with ResNet features instead of video frames enables us to train several backend configurations on a single GPU using up to 100 epochs, each within 1-2 days. Note that pre-trained ResNet networks in Torch7 and PyTorch and the code for training and evaluating them are available\footnote{https://github.com/tstafylakis/Lipreading-ResNet}. Moreover, we found no notable gains in performance by training the networks end-to-end, as long as the ResNet features are extracted with a network trained with BiLSTM backend, such as V6.

Having fixed the main building blocks of our architectures, we perform experiments in order to improve the backend. We experiment with regularization methods (dropouts and batch normalization), number of BiLSTM layers, as well as with using the last time step of BiLSTM output to aggregate temporal information, as in \cite{wand2016lipreading}. Word boundaries are included in all the experiments of this section in the form of binary indicator variables, as shown in Fig. \ref{diagram}.  

In the first experiment we include dropouts in the backend. The results are given in Table \ref{LRexp} (denoted by V11) and the network attains $13.65$\% MCR. For the second experiment (denoted by V12) we add batch normalization without dropouts to the backend. The MCR the network attained is $14.52$\%, showing the batch normalization alone is less effective compared to dropouts. In the next configuration we use both dropouts and batch normalization, but a single BiLSTM layer. The network, denoted by V13, attains $13.04$\% MCR, showing that combining dropouts with batch normalization yields good results even with a single BiLSTM layer. In the next configuration (denoted by V14) we use the last time step to aggregate temporal information instead of average pooling. By using this method (proposed e.g. in \cite{wand2016lipreading}) and combining it with dropouts, batch normalization and a 2-layer BiLSTM, the network attains $12.15$\% error rate. The final architecture we examine is the one shown in Fig. \ref{diagram} and it is denoted by V15. Its sole difference from V14 is the use of average pooling to aggregate information across time steps. Its attains $11.92$\% MCR, which is the best visual-only results we attain and corresponds to half the MCR attained by the baseline system (Table \ref{BL}). We should also mention that we experimented with the typical stacking approach of BiLSTM. In this case, the outputs of the first BiLSTM are concatenated and used as input to the second BiLSTM. The network failed to attain good results (error rates above $20$\%), despite our efforts to tune parameters such as learning rate and dropout probabilities.     

\begin{table}[!htbp]
\caption{Lipreading results for various network configurations. Abbreviations: \#L: number of BiLSTM layers, DO: use of dropouts at the backend, BN: use of batch normalization after average pooling, TA: temporal aggregation using average pooling (A) or using the last time step (L).}
\centering
\begin{tabular}{ c c c c c c } 
\hline
\bf{Net} & \bf{\#L} & \bf{DO} & \bf{BN} & \bf{TA} & \bf{MCR (\%)} \\ [0.5ex] 
\Xhline{2\arrayrulewidth}
V11 & 2  &  \checkmark &  & A & 13.65\\ 
\hline 
V12  & 2  &  & \checkmark  & A & 14.52 \\
\hline 
V13  & 1 & \checkmark & \checkmark & A &13.04 \\ 
\hline
V14 & 2 & \checkmark & \checkmark & L & 12.15\\ 
\hline
V15  & 2 & \checkmark & \checkmark & A & 11.92 \\ 
\hline
\end{tabular}
\vspace{1mm}
\label{LRexp}
\end{table}

\subsection{Experimentation on the use of word boundaries}
How effective is the proposed method of using word boundaries as indicators variables and how it compares to other ways of using them? To address this question, we first retrain the backend without passing the word boundaries. The system yields 15.7\% MCR, as Table \ref{LRWB} shows. This result is interesting in two ways; first of all, it is substantially inferior to the one attained by V15, demonstrating the utility of using word boundaries. Secondly, it is clearly superior to the current state-of-the-art (17.03\% MCR, Table \ref{BL}) demonstrating the gain in performance attained by the improved architecture. We also compare our method with the conventional approach of removing those frames lying out of word boundaries. In this case, the backend is trained with sequences of ResNet features of variable length. The MCR attained by this approach is 14.94\% (V17, Table \ref{LRWB}) which corresponds to a modest absolute improvement of 0.8\% over a network that does not use word boundaries (i.e. V16). 

\begin{table}[!htbp]
\caption{Lipreading results on the utility of word boundaries.}
\centering
\begin{tabular}{ c c c } 
\hline
\bf{Net} & \bf{Word Boundaries} &  \bf{MCR (\%)} \\ [0.5ex] 
\Xhline{2\arrayrulewidth}
V15 & Indicator variables & 11.92 \\ 
\hline 
V16 & Not used & 15.67 \\
\hline 
V17 & Frame removal & 14.94 \\ 
\hline
\end{tabular}
\vspace{1mm}
\label{LRWB}
\end{table}

Several hypotheses can be postulated from this set of experiments. First of all, LSTMs have a powerful gating mechanism and the technique of passing the word boundaries as an additional feature permits the LSTM to make use of them. Furthermore, LSTMs are capable of modeling long-range temporal dependencies, such as those related to the characteristics of the particular utterance (e.g. speaker, pose, and other utterance-level characteristics). Hence, the out-of-boundaries frames can be used by the BiLSTM backend to model such dependencies (by accumulating relevant information in the cell), and apply essentially a fast and coarse adaptation to the input ResNet features of the particular utterance, before the target-word frames arrive. Finally, the out-of-boundaries frames carry information about the linguistic context of the target word, which might be useful for identifying the target word. A further set of experiments on the utility of out-of-boundaries frames is presented in Section \ref{Sect:EA}.

The results obtained by V17 (i.e. using frame removal) can be compared to those reported in \cite{chung2018learning} using frame removal, LSTMs and a VGG-M frontend. As Table \ref{BL} shows, their architecture attains 28.50\% MCR, which is about twice the MCR obtained by the V17. This drastic improvement in performance can be attributed to the 3D-ResNet, as the other ingredients of the two networks as very similar.

For convenience, we list all the visual-only networks we examined and the corresponding MCRs in Table \ref{tbl.allvisual}.

\section{Audio and Audiovisual Experiments}
\label{Sub:AVE}
In this section, we provide experimental results attained by incorporating the audio modality in the architecture. In the first part of the section we provide details about the experimental set-up and the use of noise to make the audiovisual problem more challenging. After some first experiments where the audio modality is used alone, we explore the proposed audiovisual architectural variants and report their performance under different noisy conditions.
\subsection{Audio Experiments}
\subsubsection{Noise database and multi-condition training}
LRW is not a noise-free database as it is composed of real-world recordings from BBC-TV. On the other hand, an additional noise database is required in order to perform controlled experiments under different noisy environments and noise levels. To this end, we make use of the publicly available Diverse Environments Multichannel Acoustic Noise Database (DEMAND \citep{DEMAND}). The database contains 18 different types of background noise (noisy environments), which are partitioned into 6 categories: domestic, nature, office, public, street and transportation. Each type is recorded using a 16-channel microphone array, resulting in 16 different audio files for each of the 18 environments (i.e. 288 audio files overall). 

We use the first 4 categories for training, while testing is performed on all 6 categories. Each audio file is randomly chosen from a uniform distribution, and the same holds for the time interval from which noise is extracted. Moreover, to further enrich the noisy environments we include combinations of noises, namely noise mixtures composed of $N_n = \{0, 1, 2, 3\}$ audio files, where $N_n$ is sampled from a uniform distribution, and $N_n = 0$ corresponds to recordings without additive noise. The noise level of all noise types is uniformly distributed between -12dB to 22dB SNR, including the noise mixtures. For testing, we use fixed noisy audio files, so that each network is scored against identical test sets. Noise is sampled and added following a similar procedure with which we train the network, with two differences: (a) We do not consider mixtures of noises, and (b) we create one noisy set from all the test set of LRW, each corresponding to a predefined noise level: -10dB to 20dB SNR with 5dB step as well as on speech without additive noise. The addition of the noise takes place on the waveform domain. 

\subsubsection{Audio results without additive noise}
In Table \ref{res:clean} we present results on LRW recordings without additive noise. We test again the use of word boundaries, either as indicator variables (A3 and A4), or for removing out-of-boundaries frames (A2). We observe again the superiority of the former approach over the latter. Moreover, applying frame removal hurts recognition accuracy even when compared to not using word boundaries at all (A1). 

We moreover examine an alternative approach for performing subsampling, where we keep every odd output of the first two LSTMs instead of concatenating odd and even outputs. The network, denoted by A3, yields a slight increase in MCR compared to A4, indicating the superiority of pBiLSTM approach over simpler subsampling approaches. Finally, it is also worth examining whether the reduction of the frame rate causes any degradation in performance. To this end, we train an audio-only network (denoted by A5) which does not apply any frame rate reduction. It is similar to A4 but uses standard BiLSTMs instead of pBiLSTMs. Interestingly, we observe a slight degradation compared to A4, which shows that the use of pBiLSTMs is not harmful for the accuracy obtained by the audio component. 

\begin{table}[!htbp]
\caption{Audio results on recordings without additive noise. WB: Word Boundaries, FR: Frame Rate at the output of the second pyramidal BiLSTM. The results show that the combination of pyramidal BiLSTMs with WB used as indicator variable yields the best performance.}
\centering
\begin{tabular}{ c c c c c} 
\hline
\bf{Net} & \bf{Fr. Concat.} & \bf{WB} & \bf{FR} & \bf{MCR (\%)}\\ [0.5ex] 
\Xhline{2\arrayrulewidth}
A1 & \checkmark  & Not used & 25fps & 2.04 \\
\hline
A2 & \checkmark  & Fr. removal & 25fps & 2.08\\ 
\hline 
A3 &  & Indicat. Var. & 25fps & 1.52 \\ 
\hline
A4 & \checkmark & Indicat. Var. & 25fps & 1.38 \\ 
\hline
A5 & & Indicat. Var. & 100fps & 1.45 \\ 
\hline
\end{tabular}
\vspace{1mm}
\label{res:clean}
\end{table}

\subsection{Audiovisual experiments}
\label{AVexper}
In this section we present the results using audiovisual architectures and we compare them to our best audio-only architecture (A4). We perform experiments with 4 audiovisual networks; the first (denoted by AV1) is a intermediate integration network, the second (denoted by AV2) has the same architecture with AV1 but it is trained with multimodal training (i.e. by randomly dropping audio and video modalities as well as word boundaries, a technique inspired by \cite{chung2017lipsent}), the third (denoted by AV3) is the same as AV2 but without word boundaries, and finally the fourth (AV4) is a late integration system of our best performing audio (A4) and video (V15) networks (i.e. with word boundaries). In AV4, the weights of the video and audio networks are $\gamma$ and $1- \gamma$, respectively, where $\gamma=0.40$.  

\begin{table}[!htbp]
\caption{Audio and audiovisual results on noisy recordings. The noise level is between -10dB and 20dB and the DEMAND noise database is used. The results show (a) the vast increase in performance attained by incorporating visual information, (b) the effectiveness of multimodal training, and (c) the gap in performance between intermediate and late integration.}
\centering
\begin{tabular}{ c c c c c} 
\hline
\bf{Net} & \bf{Integration} & \bf{Multimodal} & \bf{WB} &  \bf{MCR (\%)} \\ [0.5ex] 
\Xhline{2\arrayrulewidth}
A4 & - & - & \checkmark & 6.90 \\ 
\hline
AV1 & Intermediate & & \checkmark & 2.79 \\ 
\hline 
AV2 & Intermediate & \checkmark & \checkmark & 2.20  \\
\hline 
AV3 & Intermediate & \checkmark & & 3.17 \\
\hline 
AV4 & Late &  & \checkmark & 1.90 \\ 
\hline
\end{tabular}
\vspace{1mm}
\label{res:noisy}
\end{table}

The results in Table \ref{res:noisy} demonstrate the huge improvement attained by including the visual component in noisy environments. Relative improvement equal to 72.4\% is attained by using late integration over the audio-only network. Moreover, in intermediate integration we observe the large improvement from applying multimodal training, suggested in \cite{chung2017lipsent}. Clearly, though, late integration performs better than intermediate, which is in line with other works on audiovisual integration, such as \cite{mroueh2015deep}. We conclude that more effort is required to make intermediate integration performing better than late, and methods such as multimodal training are very helpful towards this direction. Finally, AV3 is useful (a) for quantifying the contribution of word boundaries, yielding about 30\% relative improvement in terms of MCR, and (b) for comparing our architecture with other recently proposed audiovisual architectures that do not make use of word boundaries, such as \cite{petridis2018end}.

\section{Error analysis}
\label{Sect:EA}
In this section we present a detailed analysis of the results attained by our best architectures. More specifically, we break down the errors by noise category and SNR, we demonstrate how well each architecture performs on difficult word pairs and we examine to which extent the linguistic context of the target word is helping towards higher recognition accuracy. 

\subsection{Error rates for each noise level and category}

\begin{figure}[!htbp]
\centering
%
%
\definecolor{mycolor1}{rgb}{0.07843,0.16863,0.54902}%
\definecolor{mycolor2}{rgb}{0.00000,0.49804,0.00000}%
\definecolor{mycolor3}{rgb}{0.00000,0.44706,0.74118}%
\definecolor{mycolor4}{rgb}{0.47059,0.30588,0.44706}%
\definecolor{mycolor5}{rgb}{0.40000,0.00000,0.00000}%
\begin{tikzpicture}[font=\scriptsize]

\begin{axis}[%
width=2.85in,
height=2.85in,
at={(0.924in,0.687in)},
scale only axis,
xmin=-10,
xmax=25,
xlabel style={font=\color{white!15!black},yshift=+1ex},
xlabel={Signal-to-Noise Ratio (dB)},
ymin=0,
ymax=20,
ylabel style={font=\color{white!15!black},yshift=-3ex},
ylabel={MCR (\%)},
axis background/.style={fill=white},
legend style={legend cell align=left, align=left, draw=white!15!black}
]
\addplot [color=mycolor1, line width=0.7pt, mark=*, mark options={solid, mycolor1}]
  table[row sep=crcr]{%
-10	19.91\\
-5	12.32\\
0	6.88\\
5	3.62\\
10	2.29\\
15	1.73\\
20	1.54\\
25	1.51\\
};
\addlegendentry{A4}

\addplot [color=mycolor2, dashed, line width=1.0pt, mark=square, mark options={solid, mycolor2}]
  table[row sep=crcr]{%
-10	7.13\\
-5	4.13\\
0	2.60\\
5	1.79\\
10	1.43\\
15	1.26\\
20	1.20\\
25	1.20\\
};
\addlegendentry{AV1}

\addplot [color=mycolor3, dotted, line width=1.0pt, mark=o, mark options={solid, mycolor3}]
  table[row sep=crcr]{%
-10	5.12\\
-5	3.20\\
0	2.10\\
5	1.55\\
10	1.26\\
15	1.12\\
20	1.06\\
25	1.05\\
};
\addlegendentry{AV2}

\addplot [color=mycolor4, dashdotted, line width=1.0pt, mark=diamond, mark options={solid, mycolor4}]
  table[row sep=crcr]{%
-10	6.60\\
-5	4.44\\
0	3.11\\
5	2.41\\
10	2.01\\
15	1.85\\
20	1.76\\
25	1.77\\
};
\addlegendentry{AV3}

\addplot [color=mycolor5, solid, line width=1.0pt, mark=diamond, mark options={solid, mycolor5}]
  table[row sep=crcr]{%
-10	4.43\\
-5	2.72\\
0	1.83\\
5	1.31\\
10	1.08\\
15	1.00\\
20	0.95\\
25	0.96\\
};
\addlegendentry{AV4}

\end{axis}
\end{tikzpicture}%
\vspace{-2mm}\caption{MCR for audio and three audiovisual networks using DEMAND noise database (25dB correspond to speech without additive noise).}
\label{fig_DEMAND}
\end{figure}
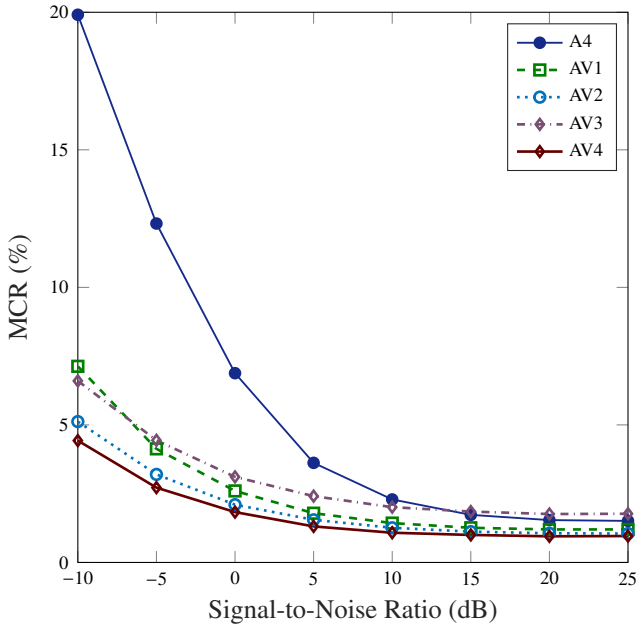

In Section \ref{AVexper} we reported results averaged over 7 different noise levels and 6 different categories. We provide here a more detailed analysis, by breaking down the error rates by noise category and noise level. In Fig. \ref{fig_DEMAND} we present MCR attained by 5 architectures, where each MCR is derived as the average over all noise categories (i.e. 6 $\times$ 25000 = 150000 videos). Although the relative improvement attained by audiovisual architectures is larger under extreme noisy conditions (-10dB to -5dB), notable improvements are attained even in higher SNRs (0dB to 20dB) as well as on speech without additive noise. Moreover, by comparing the MCR of our best visual-only network (V15) with our best audio-only network (A4) we conclude that the performance of the former is equivalent to that of the latter with $\textrm{SNR} \approx -4.5$ additive noise.

\setvariable\figwidth{0.45\textwidth}
\setvariable\figheight{0.45\textwidth}

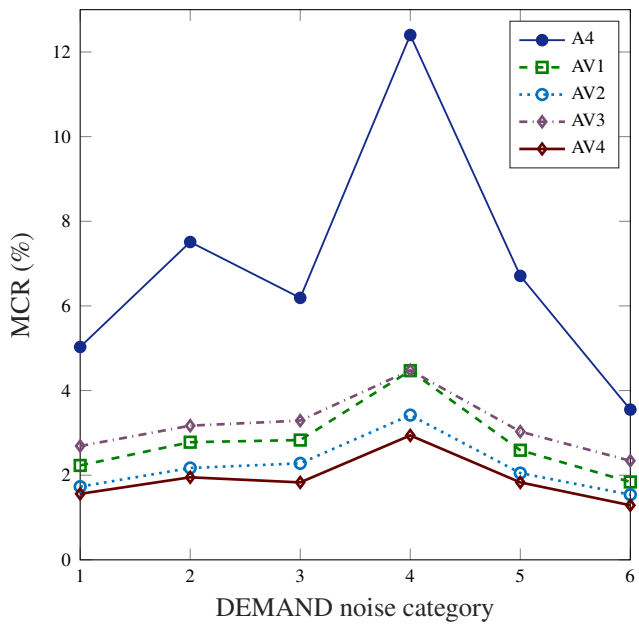
\begin{figure}[!htbp]
\centering
%
%
\definecolor{mycolor1}{rgb}{0.07843,0.16863,0.54902}%
\definecolor{mycolor2}{rgb}{0.00000,0.49804,0.00000}%
\definecolor{mycolor3}{rgb}{0.00000,0.44706,0.74118}%
\definecolor{mycolor4}{rgb}{0.47059,0.30588,0.44706}%
\definecolor{mycolor5}{rgb}{0.40000,0.00000,0.00000}%
\begin{tikzpicture}[font=\scriptsize]

\begin{axis}[%
width=2.85in,
height=2.85in,
at={(0.924in,0.687in)},
scale only axis,
xmin=1,
xmax=6,
xlabel style={font=\color{white!15!black},yshift=+1ex},
xlabel={DEMAND noise category},
ymin=0,
ymax=13,
ylabel style={font=\color{white!15!black},yshift=-3ex},
ylabel={MCR (\%)},
axis background/.style={fill=white},
legend style={legend cell align=left, align=left, draw=white!15!black}
]
\addplot [color=mycolor1, line width=0.7pt, mark=*, mark options={solid, mycolor1}]
  table[row sep=crcr]{%
1	5.03\\
2	7.51\\
3	6.19\\
4	12.40\\
5	6.71\\
6	3.55\\
};
\addlegendentry{A4}

\addplot [color=mycolor2, dashed, line width=1.0pt, mark=square, mark options={solid, mycolor2}]
  table[row sep=crcr]{%
1	2.23\\
2	2.78\\
3	2.83\\
4	4.47\\
5	2.59\\
6	1.84\\
};
\addlegendentry{AV1}

\addplot [color=mycolor3, dotted, line width=1.0pt, mark=o, mark options={solid, mycolor3}]
  table[row sep=crcr]{%
1	1.73\\
2	2.17\\
3	2.28\\
4	3.42\\
5	2.05\\
6	1.54\\
};
\addlegendentry{AV2}

\addplot [color=mycolor4, dashdotted, line width=1.0pt, mark=diamond, mark options={solid, mycolor4}]
  table[row sep=crcr]{%
1	2.69\\
2	3.17\\
3	3.29\\
4	4.48\\
5	3.03\\
6	2.34\\
};
\addlegendentry{AV3}

\addplot [color=mycolor5, solid, line width=1.0pt, mark=diamond, mark options={solid, mycolor5}]
  table[row sep=crcr]{%
1	1.56\\
2	1.95\\
3	1.83\\
4	2.94\\
5	1.83\\
6	1.29\\
};
\addlegendentry{AV4}

\end{axis}
\end{tikzpicture}%
\vspace{-2mm}\caption{MCR on the 6 noise categories of DEMAND (Domestic, Nature, Office, Public, Street and Transportation) averaged across all 7 noise SNR levels.}
\label{fig_BABBLE}
\end{figure}

Similarly, in Fig. \ref{fig_BABBLE} we report results for each noise category of the DEMAND noise database. The results are averaged across all 7 noise levels. We observe a high variability in the error rates between noise categories. The category ``public'' is the most challenging (also known as babble noise), probably due to the high spectral overlap between noise and clean speech. Interestingly, ``street'' and ``transportation'' do not yield worse results than ``domestic'', ``nature'' or  ``office'', despite the fact that the former two are not included in training.

\subsection{Challenging Word Pairs}
As there are many word pairs in LRW with similar phonetic content, it is interesting to examine the error rates attained by each architecture on such pairs. We select the most challenging pairs based on the performance of the audio-only network (A4) at 10dB and noise category ``public'' (babble noise). The list of word pairs with the highest confusion is given in Table \ref{Pairs}, together with the corresponding number of occurrences (recall there are 50 videos for each target word). The number of errors made by the proposed lipreading network (V15) on the same set of pairs is also given. Finally, we report the number of errors made by the audiovisual systems AV2 and AV4.

Table \ref{Pairs} provides some insight regarding the way visual information assists audio-only word recognition. For instance, visual representation of phoneme /b/ in the word BORDER enhances its discriminability from the word ORDER. The words SERIES and SERIOUS have lower mutual confusion in the audiovisual setting, due to the visual information in phoneme /ah/. Similarly, visual differences between phonemes /r/ and /m/ assist discrimination between words RIGHT and MIGHT. As a final example, visual information in phoneme /f/ helps to discriminate between words STAFF and START. For a thorough analysis of mappings between phonemes and visemes (i.e. visual analogues of phonemes) in lipreading we refer to \cite{bear2014phoneme}. 

\subsection{Word recognition using only the context}
As a final experiment, we aim to assess the extent to which the linguistic context helps towards recognizing the target words. The experiments on the various ways word boundaries may be utilized ($3\%$ difference in MCR between V15 and V17, Table \ref{LRWB}) indicate that out-of-boundaries frames carry useful information for word identification. However, the way the network utilizes this information is unclear. One possibility is that the backend is using out-of-boundaries frames for a coarse utterance and speaker adaptation, while a second one is that it models the linguistic context of the target word. Hence, a question worth-posing is whether or not it is possible to recognize the target words given only its context. 

\begin{table}[!h]
\caption{Word pairs exhibiting the highest confusion when the audio-only network is used (A4), and how visual information reduces the error rates, using either intermediate (AV2) or late integration (AV4, $\gamma = 0.40$). The number of test examples is 50 per word and the SNR is 10dB (DEMAND dataset).}
\centering
\begin{tabular}{ c c c c c c c c} 
\hline
 \multicolumn{2}{c}{\bf{Word Pairs}} & \multicolumn{4}{c}{\bf{\#Errors/Net}} \\
\hline
\bf{Target} & \bf{Estimated} & \bf{A4} & \bf{V15} & \bf{AV2} & \bf{AV4}\\ [0.5ex] 
\Xhline{2\arrayrulewidth}
WHETHER & WEATHER & 10 & 2 & 3 & 5 \\	
\hline	
THERE & THEIR & 8 & 6 & 8 & 8 \\	
\hline	
SPENT & SPEND & 8 & 9 & 5 & 6 \\	
\hline	
THEIR & THERE & 6 & 6 & 7 & 6 \\	
\hline	
HAPPENED & HAPPEN & 5 & 4 & 5 & 3 \\	
\hline	
WEATHER & WHETHER & 5 & 2 & 2 & 2 \\	
\hline	
STATE & STATES & 5 & 0 & 2 & 2 \\	
\hline	
CALLED & COURT & 5 & 6 & 3 & 3 \\	
\hline	
TAKING & TAKEN & 4 & 5 & 3 & 3 \\	
\hline	
SERIES & SERIOUS & 4 & 7 & 4 & 2 \\	
\hline	
ORDER & BORDER & 4 & 0 & 0 & 0 \\	
\hline	
AMERICAN & AMERICA & 4 & 10 & 3 & 3 \\	
\hline	
STAGE & STATE & 3 & 1 & 1 & 3 \\	
\hline	
STAFF & START & 3 & 3 & 0 & 1 \\	
\hline	
SCHOOL & SCHOOLS & 3 & 1 & 3 & 3 \\	
\hline	
THROUGH & THREE & 3 & 2 & 1 & 1 \\	
\hline	
SERIOUS & SERIES & 3 & 4 & 2 & 0 \\	
\hline	
MONTHS & MONTH & 3 & 2 & 1 & 2 \\	
\hline	
SPEND & SPENT & 3 & 13 & 5 & 4 \\	
\hline	
ELECTION & ACTION & 3 & 1 & 3 & 2 \\	
\hline	
THINK & THING & 3 & 5 & 4 & 5 \\	
\hline	
HAPPEN & HAPPENED & 3 & 2 & 2 & 2 \\	
\hline	
RIGHT & MIGHT & 3 & 0 & 0 & 1 \\	
\hline	
START & STAFF & 2 & 0 & 0 & 1 \\
\hline
\end{tabular}
\label{Pairs}
\end{table} 

To this end, we train a lipreading system using only out-of-boundaries frames. The network (which we denote by V18) is the dual of V17 in the sense that the within-boundaries frames are now the ones being removed. The performance of V18 is $88.86\%$ MCR, which is significantly better compared to random selection, i.e. $1 - 1/500 = 99.80\%$ MCR. Moreover, we observe certain words on which the network performs reasonably well ($\leq 50\%$ MCR), listed in Table \ref{Context}. A common characteristic of most of these words is that they are frequently preceded or succeeded by few others, such as ``David CAMERON'', ``Northern IRELAND'', ``PRIME minister'', i.e. the probability of observing them given their linguistic context is relatively high. We conclude that the network is using all available frames to create an implicit language model which helps towards distinguishing between two or more target words having similar phonetic/visemic content. This property can at least partly explain the gains in performance attained by keeping out-of-boundaries frames.

Finally, the ability of the proposed architecture to discriminate even between homophones (WHETHER vs. WEATHER) should also be linked to the use of words as main recognition units (rather than visemes, phonemes and other sub-word units). The proposed architecture, when presented with a large number of training examples seems capable of capturing certain differences in the way homophones are expressed within phrases, which cannot be captured by systems that use visemes as main recognition units (since the two sequences of visemes are identical), unless a language model is applied.

\begin{table}[!htbp]
\caption{15 words which can be recognized fairly well using only out-of-boundaries frames ($\leq 50\%$ MCR). V18: Lipreading network trained and evaluated using only out-of-boundaries frames. V15: Our best lipreading system that uses all frames.}
\centering
\begin{tabular}{ c c c} 
\hline
 \multicolumn{3}{c}{\bf{MCR (\%)}} \\
\hline
\bf{Word} & \bf{V18} & \bf{V15} \\ [0.5ex] 
\Xhline{2\arrayrulewidth}
CAMERON & 20 & 2 \\	
\hline	
IRELAND & 24 & 2 \\	
\hline	
EDITOR & 32 & 14 \\	
\hline	
PRIME & 36 & 2 \\	
\hline	
UNION & 38 & 2 \\	
\hline	
ISLAMIC & 38 & 0 \\	
\hline	
SUNSHINE & 42 & 0 \\	
\hline	
NORTHERN & 44 & 4 \\	
\hline	
MINISTER & 44 & 2 \\	
\hline	
GEORGE & 44 & 24 \\	
\hline	
SIDES & 44 & 20 \\	
\hline	
AFFAIRS & 44 & 14 \\	
\hline	
AFTERNOON & 46 & 4 \\	
\hline	
DAVID & 48 & 8 \\	
\hline	
WINDS & 50 & 2 \\	
\hline	
\end{tabular}
\vspace{1mm}
\label{Context}
\end{table} 

\section{Conclusions and further applications}
\label{Sect:Conc}
In this paper we introduced a deep architecture for audiovisual word recognition. The architecture is trained and evaluated on ``Lipreading in the wild'' (LRW), a recently released audiovisual database composed of short excerpts from BBC-TV. LRW has several appealing characteristics, such as large variability in speakers, motion and resolution, and large number of target words and training instances per word, permitting training of deep end-to-end architectures without pertained models. 

In the first part of the paper we focused on the lipreading network, where we examined several network options, regarding the use of spatiotemporal input layers in the ResNet, the size of the ResNet and the use of optical flow. The main conclusions we draw are that (a) the proposed frontend yields substantial improvement over the standard spatial ResNet, even with BiLSTM backend, (b) optical flow is not useful at least in early integration with frames, (c) there is no gain by using a 34-layer ResNet over an 18-layer one, and (d) BiLSTMs are performing much better compared to simple temporal convolutions in the backend. After fixing the architecture of the frontend, we turned our focus to the backend, where we experimented with BiLSTMs and regularization methods, as well as with method of using the word boundaries. The experiments show among others that (a) combining dropouts and batch normalization yields notably improved results, and (b) passing the word boundaries as additional binary indicator features yields substantial improvement over removing out-of-boundaries frames.  

For the audio part of the network, we proposed a deep BiLSTM network using log-spectrum as input. The backend of the audio network is identical to the visual, so that they can be merged into a single audiovisual architecture. To this end, the first two BiLSTM layers apply subsampling in their outputs, reducing the audio frame fate from 100fps to 25fps. The audio network is trained using noisy inputs, with noise samples drawn from the DEMAND database. Experiments on audio without additive noise demonstrated the capacity of the network in addressing word recognition and reaffirmed the efficacy of the method by which we make use of word boundaries. On the other hard, experiments on noisy audio showed the drastic performance degradation of audio-only ASR systems, especially when SNR $\leq$ 0dB. Furthermore, by breaking down the results by noise category, we show that babble noise (DEMAND category ``public'') is the most difficult type to mitigate.

The experiments with audiovisual architectures indicated the vast improvements attained by combining the two modalities, either in the backend (intermediate integration) or in the score-domain (late integration). Intermediate integration, especially when random droppings on each modality and word boundaries are applied during training, yields results close -yet still inferior- to those attained by late integration. For noisy audio our best intermediate integration approach (AV2) yielded 2.20\% compared to 1.90\% attained by late integration, corresponding respectively to 68\% and 72\% relative improvement over the audio network alone (A4, 6.90\% MCR). Furthermore, on speech without additive noise the two integration approaches attained relative improvement equal to 27\% and 39\% respectively, over the audio-only network. These results demonstrate the significant gains obtained by audiovisual integration even when without additive noise. 

A final set of the experiments was conducted with the aim to explain the way by which the network utilizes out-of-boundaries frames. To quantify the capacity of the architecture in predicting the target word by its linguistic context we trained and evaluated a lipreading network only on out-of-boundaries frames. The results showed that there is a fraction of target words which can be predicted fairly well by their context alone; most of these words are frequently preceded or succeeded by a small set of other words (e.g. Northern IRELAND). This experiment indicates that the network makes use of the linguistic context of the target words to increase target-word recognition accuracy and provides a further rationale for our method of using target word boundaries as additional indicator variables. 

We should finally emphasize that although the proposed networks were primarily designed to address the problem of closed-set word identification as defined by LRW, they are by no means restricted to it. The recently emerged acoustic-to-word family of models is capable of attaining state-of-the-art results in audio-only large vocabulary continuous speech recognition using words as recognition units, instead of the typical subword units such as collections of triphones \citep{li2018advancing, audhkhasi2017direct, li2017acoustic}. Visual keyword spotting is another application where words can be used as recognition units, and architectures deploying our ResNet as feature extractor (introduced in \cite{Stafy2017}) have already shown very promising results 
\citep{StafyEmbeddingsICASSP2018, StafyECCV, WordSpotting2018}. Moreover, there is evidence that the proposed training scheme (i.e. closed-set word identification on LRW with the aid of an LSTM backend) yields dynamic facial features attaining state-of-the-art performance on other tasks, which require fine-grained temporal information. Recent architectures for audiovisual speech enhancement and visual-only large vocabulary continuous speech recognition with characters as recognition units are such examples, where exceptional results are attained with our proposed ResNet features \citep{afouras2018conversation, afouras2018deep}.

\section{Acknowledgements}
This project has received funding from the European Union's Horizon 2020 research and innovation programme under the Marie Sklodowska-Curie grant agreement No 706668 (Talking Heads).

\begin{table*}[th]
  \renewcommand{\arraystretch}{1.0}
  \caption{\label{tbl.allvisual}\it{List of visual-only networks}}
  \vspace{2mm}
  \centerline
  {
    \begin{tabular}{l l c c c c c c}
      \toprule
      \bf{Net} & \bf{Input Features} & \bf{Word Boundaries} & \bf{ResNet} & \bf{Backend} & \bf{MCR} (\%) \\
      \midrule      
      V1 & Frames-2D & - & 18L-AP & TConv & 29.37 \\
      V2 & Frames-3D & - & 18L-AP & TConv & 24.71 \\
      V3 & Frames-3D & - & 34L-AP & TConv & 25.44 \\
      V4 & Frames-3D & - & 18L-FC & TConv & 22.51 \\ 
      V5 & Frames-2D & - & 18L-FC & BiLSTM(x2) & 23.28 \\
      V6 & Frames-3D & - & 18L-FC & BiLSTM(x2) & 17.01 \\ 
      V7 & Opt.Flow-2D & - & 18L-FC & TConv & 37.79 \\ 
      V8 & Opt.Flow-3D & - & 18L-FC & TConv & 34.68 \\ 
      V9 & Opt.Flow-Frames-2D & - & 18L-FC & TConv & 39.44 \\ 
      V10 & Opt.Flow-Frames-3D & - & 18L-FC & TConv & 30.28 \\  
      V11 & Frames-3D & Indicat. Var. & 18L-FC & BiLSTM(x2), w/o BN & 13.65 \\ 
      V12 & Frames-3D & Indicat. Var. & 18L-FC & BiLSTM(x2), w/o DO & 14.52 \\ 
      V13 & Frames-3D & Indicat. Var. & 18L-FC & BiLSTM(x1) & 13.04 \\ 
      V14 & Frames-3D & Indicat. Var. & 18L-FC & BiLSTM(x2), Last Frame & 12.15 \\ 
      \bf{V15} & Frames-3D & Indicat. Var. & 18L-FC & BiLSTM(x2) & \bf{11.92} \\ 
      V16 & Frames-3D & - & 18L-FC & BiLSTM(x2) & 15.67 \\ 
      V17 & Frames-3D & Fr. Removal & 18L-FC & BiLSTM(x2) & 14.94 \\ 
      V18 & Frames-3D & Within Bound. Fr. Removal & 18L-FC & BiLSTM(x2) & 88.86 \\
      \bottomrule
    \end{tabular}
  }
\end{table*}

\bibliographystyle{model2-names}
\bibliography{AV}

\end{document}